\documentclass{article}

\usepackage{times}
\usepackage{graphicx} 
\usepackage{subfigure} 

\usepackage{natbib}
\usepackage{epsfig}
\usepackage{graphicx}
\usepackage{amssymb}
\usepackage{amsmath}
\usepackage{bm,soul}

\usepackage{algorithm}
\usepackage{algorithmic}

\usepackage{hyperref}

\usepackage[accepted]{icml2017} 

\def\a{{\bf a}}
\def\b{{\bf b}}
\def\h{{\bf h}}
\def\x{{\bf x}}

\def\vh{\widehat{\bf v}}
\def\y{{\bf y}}
\def\z{{\bf z}}
\def\A{{\bf A}}
\def\D{{\bf D}}
\def\F{{\bf F}}
\def\G{{\bf G}}
\def\I{{\bf I}}
\def\P{{\bm\Pi}}
\def\R{{\bf R}}
\def\T{{\bf T}}
\def\U{{\bf U}}
\def\V{{\bf W}}
\def\W{{\bf W}}
\def\etal{{\frenchspacing\it et al.}}

\icmltitlerunning{Tunable Efficient Unitary Neural Networks (EUNN) and their application to RNNs}

\begin{document} 

\twocolumn[
\icmltitle{Tunable Efficient Unitary Neural Networks (EUNN) and their application to RNNs}

\icmlsetsymbol{equal}{*}

\begin{icmlauthorlist}
\icmlauthor{Li Jing}{equal,mit}
\icmlauthor{Yichen Shen}{equal,mit}
\icmlauthor{Tena Dubcek}{mit}
\icmlauthor{John Peurifoy}{mit}
\icmlauthor{Scott Skirlo}{mit}
\icmlauthor{Yann LeCun}{nyu}
\icmlauthor{Max Tegmark}{mit}
\icmlauthor{Marin Solja\v{c}i\'{c}}{mit}
\end{icmlauthorlist}

\icmlaffiliation{mit}{Massachusetts Institute of Technology}
\icmlaffiliation{nyu}{New York University, Facebook AI Research}

\icmlcorrespondingauthor{Li Jing}{ljing@mit.edu}
\icmlcorrespondingauthor{Yichen Shen}{ycshen@mit.edu}
\icmlkeywords{Unitary Neural Network, Recurrent Neural Network, Parametrization}

\vskip 0.3in
]

\printAffiliationsAndNotice{\icmlEqualContribution}

\begin{abstract} 
Using unitary (instead of general) matrices in artificial neural networks (ANNs) is a promising way to solve the gradient explosion/vanishing problem, as well as to enable ANNs to learn long-term correlations in the data. This approach appears particularly promising for Recurrent Neural Networks (RNNs). In this work, we present a new architecture for implementing an Efficient Unitary Neural Network (EUNNs); its main advantages can be summarized as follows. Firstly, the representation capacity of the unitary space in an EUNN is fully tunable, ranging from a subspace of SU(N) to the entire unitary space. Secondly, the computational complexity for training an EUNN is merely $\mathcal{O}(1)$ per parameter. Finally, we test the performance of EUNNs on the standard copying task, the pixel-permuted MNIST digit recognition benchmark as well as the Speech Prediction Test (TIMIT). We find that our architecture significantly outperforms both other state-of-the-art unitary RNNs and the LSTM architecture, in terms of the final performance and/or the wall-clock training speed. EUNNs are thus promising alternatives to RNNs and LSTMs for a wide variety of applications.

\end{abstract}

\section{Introduction}
Deep Neural Networks \cite{lecun2015deep} have been successful on numerous difficult machine learning tasks, including image recognition\cite{krizhevsky2012imagenet,donahue2015long}, speech recognition\cite{hinton2012deep} and natural language processing\cite{collobert2011natural,bahdanau2014neural,sutskever2014sequence}. However, deep neural networks can suffer from 
vanishing and exploding gradient problems\cite{hochreiter1991untersuchungen,bengio1994learning},
which are known to be caused by matrix eigenvalues far from unity being raised to large powers.
Because the severity of these problems grows with the the depth of a neural network, they are particularly grave for Recurrent Neural Networks (RNNs), whose recurrence can be equivalent to thousands or millions of equivalent hidden layers.

Several solutions have been proposed to solve these problems for RNNs. Long Short Term Memory (LSTM) networks \cite{hochreiter1997long}, which help RNNs contain information inside hidden layers with gates, remains one of the the most popular RNN implementations. Other recently proposed methods such as GRUs\cite{cho2014properties} and Bidirectional RNNs \cite{berglund2015bidirectional} also perform well in numerous applications. However, none of these approaches has fundamentally solved the vanishing and exploding gradient problems, and gradient clipping is often required to keep gradients in a reasonable range.

A recently proposed solution strategy is using orthogonal hidden weight matrices or their complex generalization (unitary matrices) \cite{saxe2013exact,le2015simple,arjovsky2015unitary,henaff2016orthogonal}, 
because all their eigenvalues will then have absolute values of unity, and can safely be raised to large powers. This has been shown to help both when weight matrices are initialized to be unitary
\cite{saxe2013exact,le2015simple} and when they are kept unitary during training, either by restricting them to a more tractable matrix subspace \cite{arjovsky2015unitary} or by alternating gradient-descent steps with projections onto the unitary subspace \cite{wisdom2016full}.

In this paper, we will first present an Efficient Unitary Neural Network (EUNN) architecture that parametrizes the entire space of unitary matrices in a complete and computationally efficient way, thereby eliminating the need for time-consuming unitary subspace-projections. Our architecture has a wide range of capacity-tunability to represent subspace unitary models by fixing some of our parameters; the above-mentioned unitary subspace models correspond to special cases of our architecture. We also implemented an EUNN with an earlier introduced FFT-like architecture which efficiently approximates the unitary space with minimum number of required parameters\cite{Mathieu2014Fast}.

We then benchmark EUNN's performance on both simulated and real tasks: the standard copying task, the pixel-permuted MNIST task, and speech prediction with the TIMIT dataset \cite{garofolo1993darpa}. 
We show that our EUNN algorithm with an $\mathcal{O}(N)$ hidden layer size can compute up to the entire $N\times N$ gradient matrix using $\mathcal{O}(1)$ computational steps and memory access per parameter. This is superior to the $\mathcal{O}(N)$ computational complexity of the existing training method for a full-space unitary network \cite{wisdom2016full} and $\mathcal{O}(\log N)$ more efficient than the subspace Unitary RNN\cite{arjovsky2015unitary}.

\section{Background}
\subsection{Basic Recurrent Neural Networks}

A recurrent neural network takes an input sequence and uses the current hidden state to generate a new hidden state during each step, memorizing past information in the hidden layer. We first review the basic RNN architecture.

Consider an RNN updated at regular time intervals $t=1,2,...$ whose input is the sequence of vectors $\x^{(t)}$ whose hidden layer $\h^{(t)}$ is updated according to the following rule: 
\begin{equation}
\h^{(t)}=\sigma(\U\x^{(t)}+\bm{W}\h^{(t-1)}),
\end{equation}
where $\sigma$ is the nonlinear activation function. 
The output is generated by
\begin{equation}
\y^{(t)}=\V\h^{(t)}+\b,
\end{equation}
where $\b$ is the bias vector for the hidden-to-output layer.
For $t=0$, the hidden layer $\mathbf{h^{(0)}}$ can be initialized to some special vector or set as a trainable variable.
For convenience of notation, we define $\z^{(t)} =\U\x^{(t)}+\W\h^{(t-1)}$ so that $\h^{(t)}=\sigma({\z^{(t)}})$.

\subsection{The Vanishing and Exploding Gradient Problems}

When training the neural network to minimize a cost function $C$ that depends on a parameter vector $\a$, the gradient descent method updates this vector to 
$\a-\lambda\frac{\partial C}{\partial\a}$, where $\lambda$ is a fixed learning rate
and $\frac{\partial C}{\partial\a}\equiv\nabla C$.
For an RNN, the vanishing or exploding gradient problem is most significant during back propagation from hidden to hidden layers, so we will only focus on the gradient for hidden layers. Training the input-to-hidden and hidden-to-output matrices is relatively trivial once the hidden-to-hidden matrix has been successfully optimized.

In order to evaluate $\frac{\partial C}{\partial W_{ij}}$, one first computes the derivative $\frac{\partial C}{\partial h^{(t)}}$ using the chain rule:
\begin{eqnarray}
\frac{\partial C}{\partial\h^{(t)}} &=& \frac{\partial C}{\partial\h^{(T)}} \frac{\partial\h^{(T)}}{\partial\h^{(t)}} \\
&=&\frac{\partial C}{\partial\h^{(T)}}\prod_{k=t}^{T-1} \frac{\partial \h^{(k+1)}}{\partial\h^{(k)}}\\
&=&\frac{\partial C}{\partial\h^{(T)}}\prod_{k=t}^{T-1} \D^{(k)} {\W},
\end{eqnarray} 
where $\D^{(k)}= {\rm diag}\{\sigma'(\U\x^{(k)}+\W\h^{(k-1)})\}$ is the Jacobian matrix of the pointwise nonlinearity.
For large times $T$, the term $\prod \W$ plays a significant role. 
As long as the eigenvalues of $\D^{(k)}$ are of order unity, then
if $\W$ has eigenvalues $\lambda_i\gg 1$, they will cause gradient explosion 
$|\frac{\partial C}{\partial\h^{(T)}}|\rightarrow \infty$,
while if $\W$ has eigenvalues $\lambda_i\ll 1$,
they can cause gradient vanishing, 
$|\frac{\partial C}{\partial\h^{(T)}}|\rightarrow 0$.
Either situation prevents the RNN from working efficiently.

\section{Unitary RNNs}

\subsection{Partial Space Unitary RNNs}

In a breakthrough paper, Arjovsky, Shah \& Bengio  \cite{arjovsky2015unitary} showed that unitary RNNs can overcome the exploding and vanishing gradient problems and perform well on long term memory tasks
if the hidden-to-hidden matrix in parametrized in the following unitary form:
\begin{equation}
\W = \D_3\T_2\mathcal{F}^{-1}\D_2\P\T_1\mathcal{F}\D_1.
\end{equation}
Here $\D_{1,2,3}$ are diagonal matrices with each element $e^{i\omega_j}, j=1,2,\cdots,n$. $\T_{1,2}$ are reflection matrices, and $\T=I-2\frac{\vh\vh^\dagger}{||\vh||^2}$, where $\vh$ is a vector with each of its entries as a parameter to be trained. $\bm{\Pi}$ is a fixed permutation matrix. $\mathcal{F}$ and $\mathcal{F}^{-1}$ are Fourier and inverse Fourier transform matrices respectively. Since each factor matrix here is unitary, the product $\W$ is also a unitary matrix.

This model uses $\mathcal{O}(N)$ parameters, which spans merely a part of the whole $\mathcal{O}(N^2)$-dimensional space of unitary $N\times N$ matrices to enable computational efficiency.
Several subsequent papers have tried to expand the space to $\mathcal{O}(N^2)$ in order to achieve better performance, as summarized below.

\subsection{Full Space Unitary RNNs}

In order to maximize the power of Unitary RNNs, it is preferable to have the option to optimize the weight matrix $\W$ over the full space of unitary matrices rather than a subspace as above. 
A straightforward method for implementing this is by simply updating $\W$ with standard back-propagation and then projecting the resulting matrix (which will typically no longer be unitary)
back onto to the space of unitary matrices. Defining $\G_{ij}\equiv\frac{\partial{C}}{\partial{W_{ij}}}$ as the gradient with respect to  $\W$, this can be implemented by 
the procedure defined by  \cite{wisdom2016full}:
\begin{eqnarray}
\A^{(t)} &\equiv&{\G^{(t)}}^{\dagger}\W^{(t)} - {\W^{(t)}}^{\dagger}\G^{(k)},\\
\W^{(t+1)} &\equiv& \left(\I+\frac{\lambda}{2}\A^{(t)}\right)^{-1}\left(\I-\frac{\lambda}{2}\A^{(t)}\right)\W^{(t)}.
\end{eqnarray}
This method shows that full space unitary networks are superior on many RNN tasks \cite{wisdom2016full}.
A key limitation is that the back-propation in this method cannot avoid $N$-dimensional matrix multiplication, incurring $\mathcal{O}(N^3)$ computational cost.

\section{Efficient Unitary Neural Network (EUNN) Architectures}

In the following, we first describe a general parametrization method able to represent arbitrary unitary matrices with up to $N^2$ degrees of freedom. We then present an efficient algorithm for this parametrization scheme, requiring only $\mathcal{O}(1)$ computational and memory access steps to obtain the gradient for each parameter. Finally, we show that our scheme performs significantly better than the above mentioned methods on a few well-known benchmarks.

\subsection{Unitary Matrix Parametrization}

\begin{figure}[h]
\centering
\includegraphics[width=3.4in]{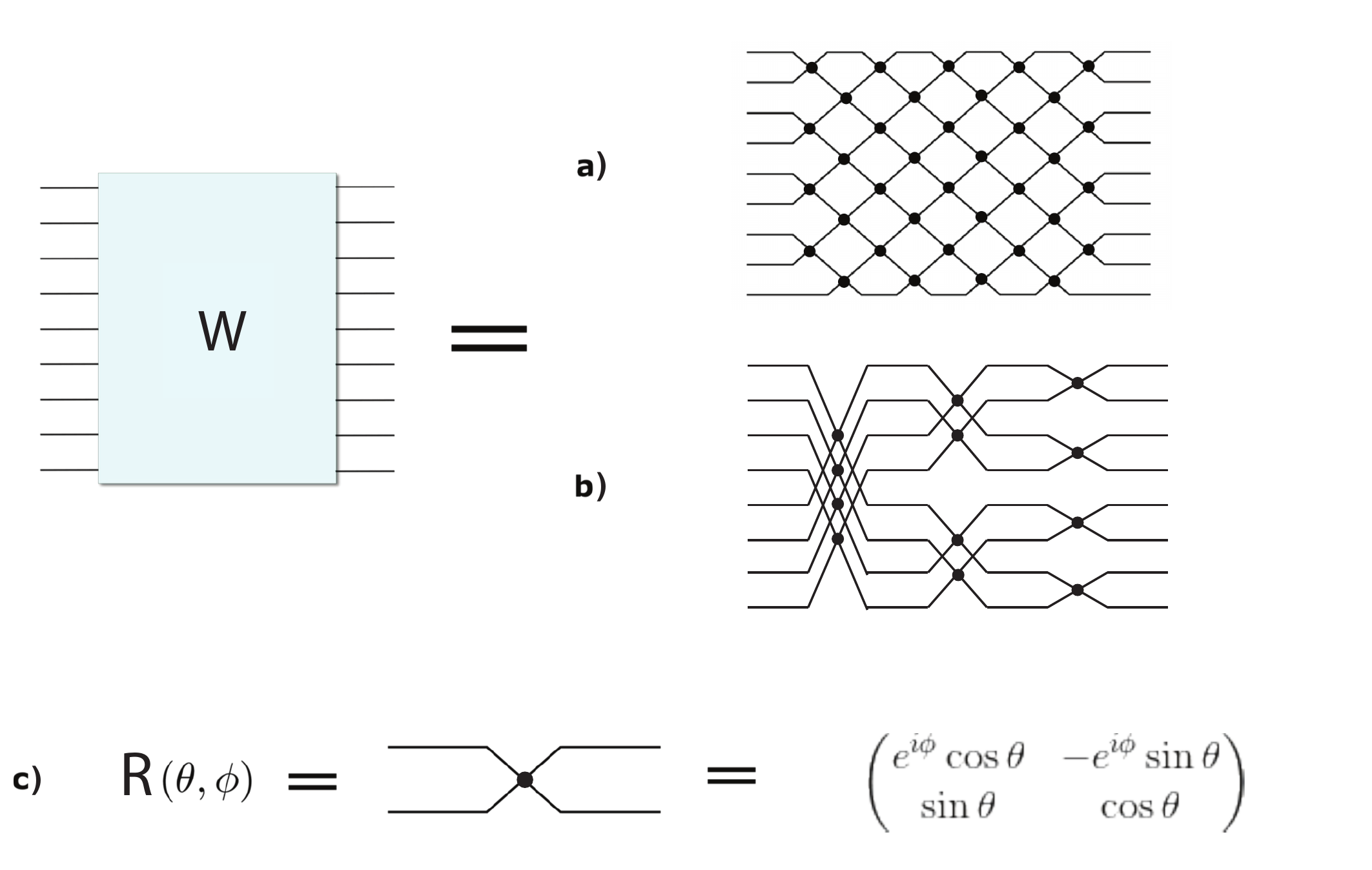}
\caption{\textbf{Unitary matrix decomposition:} An arbitrary unitary matrix $\W$ can be decomposed (a) with the square decomposition method of Clements {\protect\etal} \cite{Clements2015optimal} discussed in section 4.2; or approximated (b) by the Fast Fourier Transformation(FFT) style decomposition method \cite{Mathieu2014Fast} discussed in section 4.3. Each junction in the a) and b) graphs above represent the U(2) matrix as shown in c).}
\label{fig:decomp}
\end{figure}

Any $N\times N$ unitary matrix $\W_N$ can be represented as a product of rotation matrices $\{\R_{ij}\}$ and a diagonal matrix $\D$, 
such that $\W_N=\D\prod_{i=2}^{N}\prod_{j=1}^{i-1}\R_{ij}$,
where $\R_{ij}$ is defined as the $N$-dimensional identity matrix with the elements 
$R_{ii}$, $R_{ij}$, $R_{ji}$ and $R_{jj}$ replaced 
as follows \cite{Reck1994experimental,Clements2015optimal}: 
\begin{equation}
\begin{pmatrix}
R_{ii}&R_{ij}\\ 
R_{ji}&R_{jj}\\
\end{pmatrix}
=
\begin{pmatrix}
e^{i\phi_{ij}}\cos\theta_{ij} &-e^{i\phi_{ij}}\sin\theta_{ij}\\ 
\sin\theta_{ij}&\cos\theta_{ij}\\
\end{pmatrix}.
\end{equation}
where $\theta_{ij}$ and $\phi_{ij}$ are unique parameters corresponding to $\mathbf{R_{ij}}$. Each of these matrices performs a $U(2)$ unitary transformation on a two-dimensional subspace of the N-dimensional Hilbert space, leaving an $(N-2)$-dimensional subspace unchanged. 
In other words, a series of $U(2)$ rotations can be used to successively make all off-diagonal elements of the given $N \times N$ unitary matrix zero. This generalizes the familiar factorization of a 3D rotation matrix into 2D rotations parametrized by the three Euler angles.
To provide intuition for how this works, let us briefly describe a simple way of doing this that is similar to Gaussian elimination by finishing one column at a time. There are infinitely many alternative decomposition schemes as well; Fig.~\ref{fig:decomp} shows two that are particularly convenient to implement in software (and even in neuromorphic hardware \cite{shen2016deep}).
The unitary matrix $\W_N$ is multiplied from the right by a succession of unitary matrices $\R_{Nj}$ for $j = N-1, \cdots, 1$. Once all elements of the last row except the one on the diagonal are zero, this row will not be affected by later transformations. Since all transformations are unitary, the last column will then also contain only zeros except on the diagonal:
\begin{eqnarray}
\W_N\R_{N,N-1}\R_{N,N-2}\cdot\cdot\R_{N,1}=
\begin{pmatrix}
\W_{N-1} & 0\\ 
0 & e^{iw_N}
\end{pmatrix}
\end{eqnarray}

The effective dimensionality of the the matrix $\W$ is thus reduced to $N-1$. The same procedure can then be repeated $N-1$ times until the effective dimension of $\W$ is reduced to 1, leaving us with a diagonal matrix:\footnote{Note that Gaussian Elimination would make merely the upper triangle of a matrix vanish, requiring a subsequent series of rotations (complete Gauss-Jordan Elimination) to zero the lower triangle. We need no such subsequent series because since $\W$ is unitary:
it is easy to show that if a unitary matrix is triangular, it must be diagonal.}
\begin{equation}
\W_N\R_{N,N-1}\R_{N,N-2}\cdots\R_{i,j}\R_{i,j-1}\cdots\R_{3,1}\R_{2,1} = \D,
\end{equation}
where $\D$ is a diagonal matrix whose diagonal elements are $e^{iw_j}$, from which we can write the direct representation of $\W_N$ as

\begin{eqnarray}
\label{eq:W_N}
\W_N&=\D\R_{2,1}^{-1}\R_{3,1}^{-1}\dots\R_{N,N-2} ^{-1}\R_{N,N-1}^{-1}\nonumber\\
&=\D\R'_{2,1}\R'_{3,1}\dots\R'_{N,N-2}\R'_{N,N-1}.
\end{eqnarray}
where \begin{equation}\R'_{ij}=\R(-\theta_{ij},-\phi_{ij})=\R(\theta_{ij},\phi_{ij})^{-1}=\R_{ij}^{-1}\end{equation}This parametrization thus involves 
$N(N-1)/2$ different $\theta_{ij}$-values,
$N(N-1)/2$ different $\phi_{ij}$-values 
and $N$ different $w_i$-values, combining to $N^2$ parameters in total and spans the entire unitary space. Note we can always fix a portion of our parameters, to span only a subset of unitary space -- indeed, our benchmark test below will show that for certain tasks, full unitary space parametrization is not necessary. \footnote[2]{Our preliminary experimental tests even suggest that a
 full-capacity unitary RNN is even undesirable for some tasks.}

\subsection{Tunable space implementation}

The representation in Eq.~\ref{eq:W_N} can be made more compact by reordering and grouping specific rotational matrices, as was shown in the optical community~\cite{Reck1994experimental,Clements2015optimal} in the context of universal multiport interferometers.
For example~\cite{Clements2015optimal}, a unitary matrix can be decomposed as
\begin{eqnarray}
\W_N&=&\D
\left(\R_{1,2}^{(1)}\R_{3,4}^{(1)}\dots\R_{N/2-1,N/2}^{(1)}\right)\nonumber\\
&&\times\left(\R_{2,3}^{(2)}\R_{4,5}^{(2)}\dots\R_{N/2-2,N/2-1}^{(2)}\right)\nonumber\\
&&\times\dots\nonumber\\
&=&\D \F_{A}^{(1)} \F_B^{(2)} \dots \F_{B}^{(L)} ,
\label{Eqn:FA}
\end{eqnarray}
where every
\begin{eqnarray}
\mathbf{F}_A^{(l)} =\R_{1,2}^{(l)}\R_{3,4}^{(l)}\dots\R_{N/2-1,N/2}^{(l)}\nonumber
\end{eqnarray}
is a block diagonal matrix, with $N$ angle parameters in total, and
\begin{eqnarray}
\mathbf{F}_B^{(l)} =\R_{2,3}^{(l)}\R_{4,5}^{(l)}\dots\R_{N/2-2,N/2-1}^{(l)} \nonumber
\end{eqnarray}
with $N-1$ parameters,
as is schematically shown in Fig.~\ref{fig:decomp}a.
By choosing different values for $L$ , $\W_N$ will span a different subspace of the unitary space. Specifically,when $L=N$, $\W_N$ will span the entire unitary space. 

Following this physics-inspired scheme, we decompose our unitary hidden-to-hidden layer matrix $\W$ as
\begin{equation}
\mathbf{W} = \mathbf{D}\mathbf{F}_A^{(1)}\mathbf{F}_B^{(2)}\mathbf{F}_A^{(3)}\mathbf{F}_B^{(4)}\cdots\mathbf{F}_B^{(L)}.
\end{equation}


\subsection{FFT-style approximation}
Inspired by \cite{michael2014mathieu}, an alternative way to organize the rotation matrices is implementing an FFT-style architecture. Instead of using adjacent rotation matrices, each $\mathbf{F}$ here performs a certain distance pairwise rotations as shown in Fig.~\ref{fig:decomp}b: 
\begin{equation}
\mathbf{W} = \mathbf{D}\mathbf{F}_1\mathbf{F}_2\mathbf{F}_3\mathbf{F}_4\cdots\mathbf{F}_{\log(N)}.
\label{Eqn:F1}
\end{equation}

The rotation matrices in $\mathbf{F}_i$ are performed between pairs of coordinates 
\begin{equation}
(2pk + j, p(2k+1) + j)
\end{equation} 
where $p=\frac{N}{2^i}$, $k\in\{0,...,2^{i-1}\}$ and $j\in\{1,...,p\}$. This requires only  $\log(N)$ matrices, so there are a total of $N\log(N)/2$ rotational pairs. This is also the minimal number of rotations that can have all input coordinates interacting with each other, providing an approximation of arbitrary unitary matrices.




\subsection{Efficient implementation of rotation matrices}

To implement this decomposition efficiently in an RNN, we apply vector element-wise multiplications and permutations:
we evaluate the product $\mathbf{F}\x$ as
\begin{equation}
\mathbf{F}\mathbf{x}= \mathbf{v_1}*\mathbf{x} +\mathbf{v_2} * \mathrm{permute}(\mathbf{x})
\end{equation}
where $*$ represents element-wise multiplication, $\mathbf{F}$ refers to general rotational matrices such as $\mathbf{F}_{A/B}$ in Eq.~\ref{Eqn:FA} and $\mathbf{F}_i$ in Eq.~\ref{Eqn:F1}. For the case of the tunable-space implementation, if we want to implement $\mathbf{F}_A^{(l)}$ in Eq.~\ref{Eqn:FA}, we define $\mathbf{v}$ and the permutation as follows:
\begin{eqnarray*}
\mathbf{v_1} = (e^{i\phi_1^{(l)}}\cos\theta_1^{(l)}, \cos\theta_1^{(l)}, e^{i\phi_2^{(l)}}\cos\theta_2^{(l)}, \cos\theta_2^{(l)}, \cdots)\\
\mathbf{v_2} = (-e^{i\phi_1^{(l)}}\sin\theta_1^{(l)}, \sin\theta_1^{(l)}, -e^{i\phi_2^{(l)}}\sin\theta_2, \sin\theta_2^{(l)}, \cdots)\\
\mathrm{permute}(\mathbf{x}) = (x_2, x_1, x_4, x_3, x_6, x_5, \cdots).
\end{eqnarray*}
For the FFT-style approach, if we want to implement $\mathbf{F}_1$ in Eq ~\ref{Eqn:F1}, we define $\mathbf{v}$ and the permutation as follows:
\begin{eqnarray*}
\mathbf{v_1} = (e^{i\phi_1^{(l)}}\cos\theta_1^{(l)}, e^{i\phi_2^{(l)}}\cos\theta_2^{(l)}, \cdots, \cos\theta_1^{(l)}, \cos\theta_2^{(l)}, \cdots)\\
\mathbf{v_2} = (-e^{i\phi_1^{(l)}}\sin\theta_1^{(l)}, -e^{i\phi_2^{(l)}}\sin\theta_2, \cdots, \sin\theta_1^{(l)}, \sin\theta_2^{(l)}, \cdots)\\
\mathrm{permute}(\mathbf{x}) = (x_{\frac{n}{2}+1}, x_{\frac{n}{2}+2} \cdots x_{n}, x_1, x_2 \cdots).
\end{eqnarray*}

In general, the pseudocode for implementing operation $\mathbf{F}$ is as follows:
\begin{algorithm}[h!]
   \caption{Efficient implementation for $\mathbf{F}$ with parameter $\theta_i$ and $\phi_i$.}
   \label{alg:example}
\begin{algorithmic}[h!]
   \STATE {\bfseries Input:} input $\mathbf{x}$, size $N$; parameters $\mathbf{\theta}$ and $\mathbf{\phi}$, size $N/2$; constant permuatation index list $\mathbf{ind_1}$ and $\mathbf{ind_2}$.
   \STATE {\bfseries Output:} output $\mathbf{y}$, size $N$.
   \STATE $\mathbf{v_1}$ $\leftarrow$ concatenate($\cos\mathbf{\theta}$, $\cos\mathbf{\theta}$ * $\exp(i\mathbf{\phi})$)
   \STATE $\mathbf{v_2}$ $\leftarrow$ concatenate($\sin\mathbf{\theta}$, - $\sin\mathbf{\theta}$ * $\exp(i\mathbf{\phi})$)
   \STATE $\mathbf{v_1}$ $\leftarrow$ $\mathrm{permute}(\mathbf{v_1}, \mathbf{ind_1})$
   \STATE $\mathbf{v_2}$ $\leftarrow$ $\mathrm{permute}(\mathbf{v_2}, \mathbf{ind_1})$
   \STATE $\mathbf{y}$ $\leftarrow$ $\mathbf{v_1} * \mathbf{x} + \mathbf{v_2} * \mathrm{permute}(\mathbf{x}, \mathbf{ind_2})$

\end{algorithmic}
\end{algorithm}

Note that $\mathbf{ind_1}$ and $\mathbf{ind_2}$ are different for different $\mathbf{F}$.

From a computational complexity viewpoint, since the operations $*$ and $\mathrm{permute}$ take $\mathcal{O}(N)$ computational steps, evaluating $\mathbf{F}\mathbf{x}$ only requires $\mathcal{O}(N)$ steps. 
The product $\mathbf{D}\mathbf{x}$ is trivial, consisting of an element-wise vector multiplication. Therefore, the product $\mathbf{W}\mathbf{x}$ with the total unitary matrix $\mathbf{W}$ can be computed in only $\mathcal{O}(NL)$  steps, and only requires $\mathcal{O}(NL)$ memory access (for full-space implementation $L=N$, for FFT-style approximation gives $L=\log N$). A detailed comparison on computational complexity of the existing unitary RNN architectures is given in Table 1.

\begin{table*}[t!]
\centering
\caption{Performance comparison of four Recurrent Neural Network algorithms:
URNN \cite{arjovsky2015unitary}, PURNN \cite{wisdom2016full},
and EURNN (our algorithm).
$T$ denotes the RNN length and $N$ denotes the hidden state size. For the tunable-style EURNN, $L$ is an integer between 1 and $N$ parametrizing the unitary matrix capacity.}
\vspace{0.1in}
\begin{tabular}{cccc}
\hline
Model           & Time complexity of one & number of parameters & Transition matrix \\
				&  online gradient step & in the hidden matrix & search space \\
\hline
URNN       &  $\mathcal{O}(TN\log N)$ & $\mathcal{O}(N)$ & subspace of $\mathbf{U}(N)$ \\
PURNN  & $\mathcal{O}(TN^2+N^3)$ & $\mathcal{O}(N^2)$ & full space of $\mathbf{U}(N)$ \\
EURNN (tunable style) & $\mathcal{O}(TNL)$ & $\mathcal{O}(NL)$ & tunable space of $\mathbf{U}(N)$ \\
EURNN (FFT style) & $\mathcal{O}(TN\log N)$ & $\mathcal{O}(N\log N)$ & subspace of $\mathbf{U}(N)$ \\
\hline
\end{tabular}
\label{tab:mnist}
\end{table*}

\subsection{Nonlinearity}
We use the same nonlinearity as \cite{arjovsky2015unitary}:
\begin{equation}
(\mathrm{modReLU}(\mathbf{z}, \mathbf{b}))_i = \frac{z_i}{|z_i|}*\mathrm{ReLU}(|z_i| + b_i)
\end{equation}
where the bias vector $\mathbf{b}$ is a shared trainable parameter, and $|z_i|$ is the norm of the complex number $z_i$. 

For real number input, $\mathrm{modReLU}$ can be simplified to:
\begin{equation}
(\mathrm{modReLU}(\mathbf{z}, \mathbf{b}))_i = \mathrm{sign}(z_i)*\mathrm{ReLU}(|z_i| + b_i)
\end{equation}
where $|z_i|$ is the absolute value of the real number $z_i$. 

We empirically find that this nonlinearity function performs the best. We believe that this function possibly also serves as a forgetting filter that removes the noise using the bias threshold.

\section{Experimental tests of our method}

In this section, we compare the performance of our Efficient Unitary Recurrent Neural Network (EURNN) with 
\begin{enumerate}
\item an LSTM RNN \cite{hochreiter1997long},
\item a Partial Space URNN \cite{arjovsky2015unitary}, and
\item a Projective full-space URNN \cite{wisdom2016full}. 
\end{enumerate}

All models are implemented in both Tensorflow and Theano, available from \url{https://github.com/jingli9111/EUNN-tensorflow} and \url{https://github.com/iguanaus/EUNN-theano}.

\subsection{Copying Memory Task}
We compare these networks by applying them all to the well defined {\it Copying Memory Task} \cite{hochreiter1997long,arjovsky2015unitary,henaff2016orthogonal}. The copying task is a synthetic task that is commonly used to test the network's ability to remember information seen $T$ time steps earlier. 

Specifically, the task is defined as follows
\cite{hochreiter1997long,arjovsky2015unitary,henaff2016orthogonal}.
An alphabet consists of symbols  $\{a_i\}$, the first $n$ of which represent data, and the remaining two representing ``blank'' and ``start recall", respectively; as illustrated by the following example where $T=20$ and $M=5$: 
\begin{verbatim}
Input:  BACCA--------------------:----
Output: -------------------------BACCA
\end{verbatim}
In the above example, $n=3$ and
$\{a_i\}=\{A, B, C,-,:\}$.
The input consists of $M$ random data symbols ($M=5$ above) followed by $T-1$ blanks, the ``start recall" symbol and $M$ more blanks. The desired output consists of $M+T$ blanks followed by the data sequence.
The cost function $C$ is defined as the cross entropy of the input and output sequences, which vanishes for perfect performance.

\begin{figure}[h!]
\centering
\includegraphics[width=3.5in]{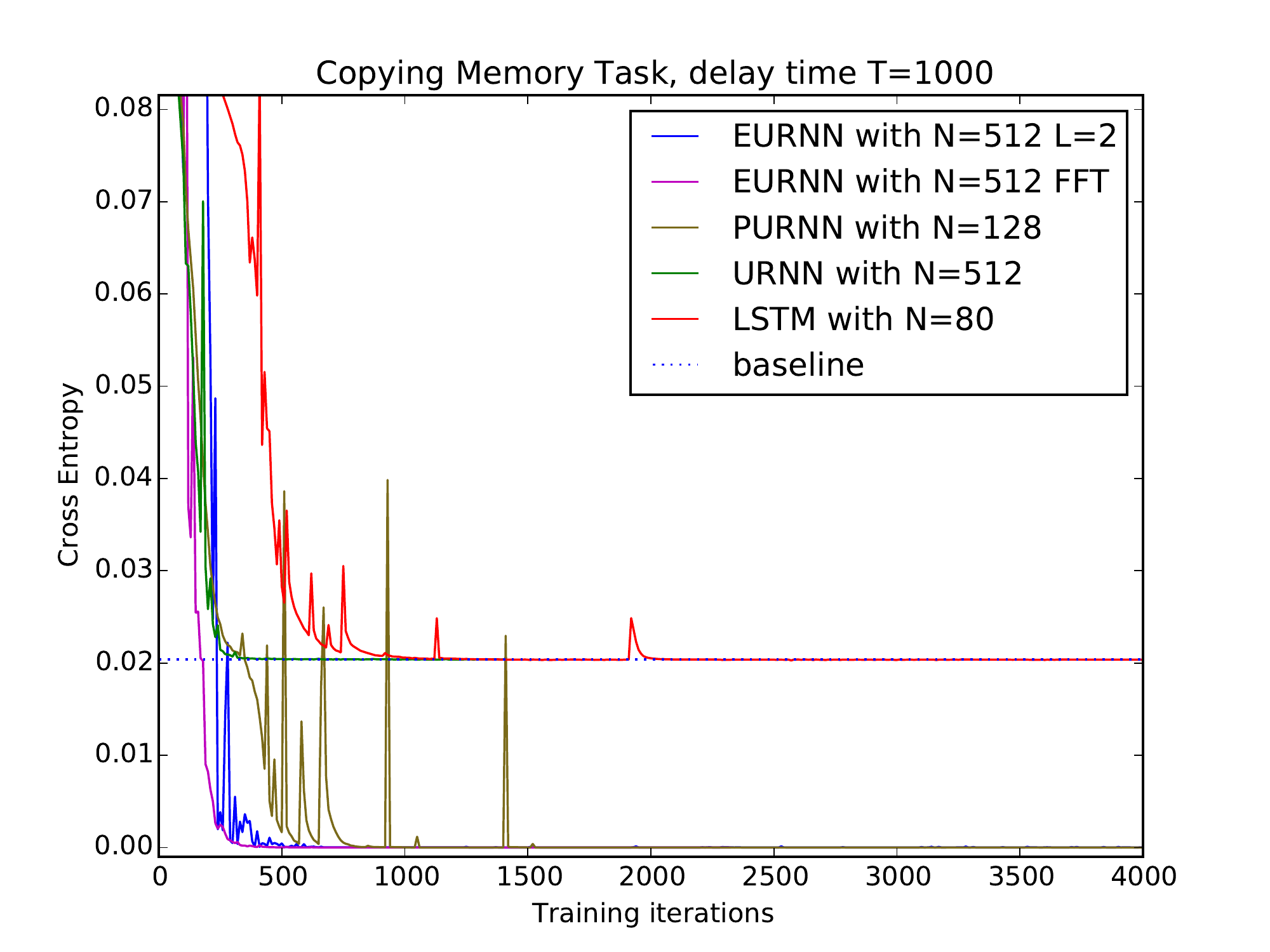}
\caption{Copying Task for $T=1000$. EURNN corresponds to our algorithm, projective URNN corresponds to algorithm presented in \cite{wisdom2016full}, URNN corresponds to the algorithm presented in 
\cite{arjovsky2015unitary}. A useful baseline performance is that of the memoryless strategy, which outputs 
$M+T$ blanks followed by $M$ random data symbols and produces a cross entropy $C=(M\log{n})/(T+2*M)$.
[Note that each iteration for PURNN takes about 32 times longer than for EURNN models, for this particular simulation, so the speed advantage is much greater than apparent in this plot.]
}
\label{fig:copying1}
\end{figure}

We use $n=8$ and input length $M=10$. The symbol for each input is represented by an $n$-dimensional one-hot vector. We trained all five RNNs for $T=1000$ with the same batch size 128 using RMSProp optimization with a learning rate of 0.001. The decay rate is set to 0.5 for EURNN, and 0.9 for all other models respectively. (Fig.~\ref{fig:copying1}). This results show that the EURNN architectures introduced in both Sec.4.2 (EURNN with N=512, selecting L=2) and Sec.4.3 (FFT-style EURNN with N=512) outperform the LSTM model (which suffers from long term memory problems and only performs well on the copy task for small time delays $T$) and all other unitary RNN models, both in-terms of learnability and in-terms of convergence rate. Note that the only other unitary RNN model that is able to beat the baseline for $T=1000$ \cite{wisdom2016full} is significantly slower than our method.

Moreover, we find that by either choosing smaller $L$ or by using the FFT-style method (so that $\mathbf{W}$ spans a smaller unitary subspace), the EURNN converges toward optimal performance significantly more efficiently (and also faster in wall clock time) than the partial \cite{arjovsky2015unitary} and projective \cite{wisdom2016full} unitary methods. The EURNN also performed more robustly. This means that a full-capacity unitary matrix is not necessary for this particular task.

\subsection{Pixel-Permuted MNIST Task}
The MNIST handwriting recognition problem is one of the classic benchmarks for quantifying the learning ability of neural networks. MNIST images are formed by a 28$\times$28 grayscale image with a target label between 0 and 9. 

To test different RNN models, we feed all pixels of the MNIST images into the RNN models in 28$\times$28 time steps, where one pixel at a time is fed in as a floating-point number. A fixed random permutation is applied to the order of input pixels. The output is the probability distribution quantifying the digit prediction. We used RMSProp with a learning rate of 0.0001 and a decay rate of 0.9, and set the batch size to 128. 

\begin{table*}[t!]
\centering
\caption{MNIST Task result. EURNN corresponds to our algorithm,  PURNN corresponds to algorithm presented in \cite{wisdom2016full}, URNN corresponds to the algorithm presented in \cite{arjovsky2015unitary}.}
\vspace{0.1in}
\begin{tabular}{ccccc}
\hline
Model           & hidden size & number of  & validation &  test \\
				&   (capacity) & parameters & accuracy & accuracy \\
\hline
LSTM					& 80 & 16k & 0.908 & 0.902  \\
URNN       & 512 & 16k & \bf{0.942} & 0.933  \\
PURNN  & 116 & 16k & 0.922 & 0.921  \\
EURNN (tunable style)     & 1024 (2) & 13.3k & 0.940 & \bf{0.937}  \\
EURNN (FFT style)   & 512 (FFT) & 9.0k & 0.928 & 0.925  \\
\hline
\end{tabular}
\label{tab:mnist}
\end{table*}

\begin{figure}[h!]
\label{fig:copying2}
\centering
\includegraphics[width=3.5in]{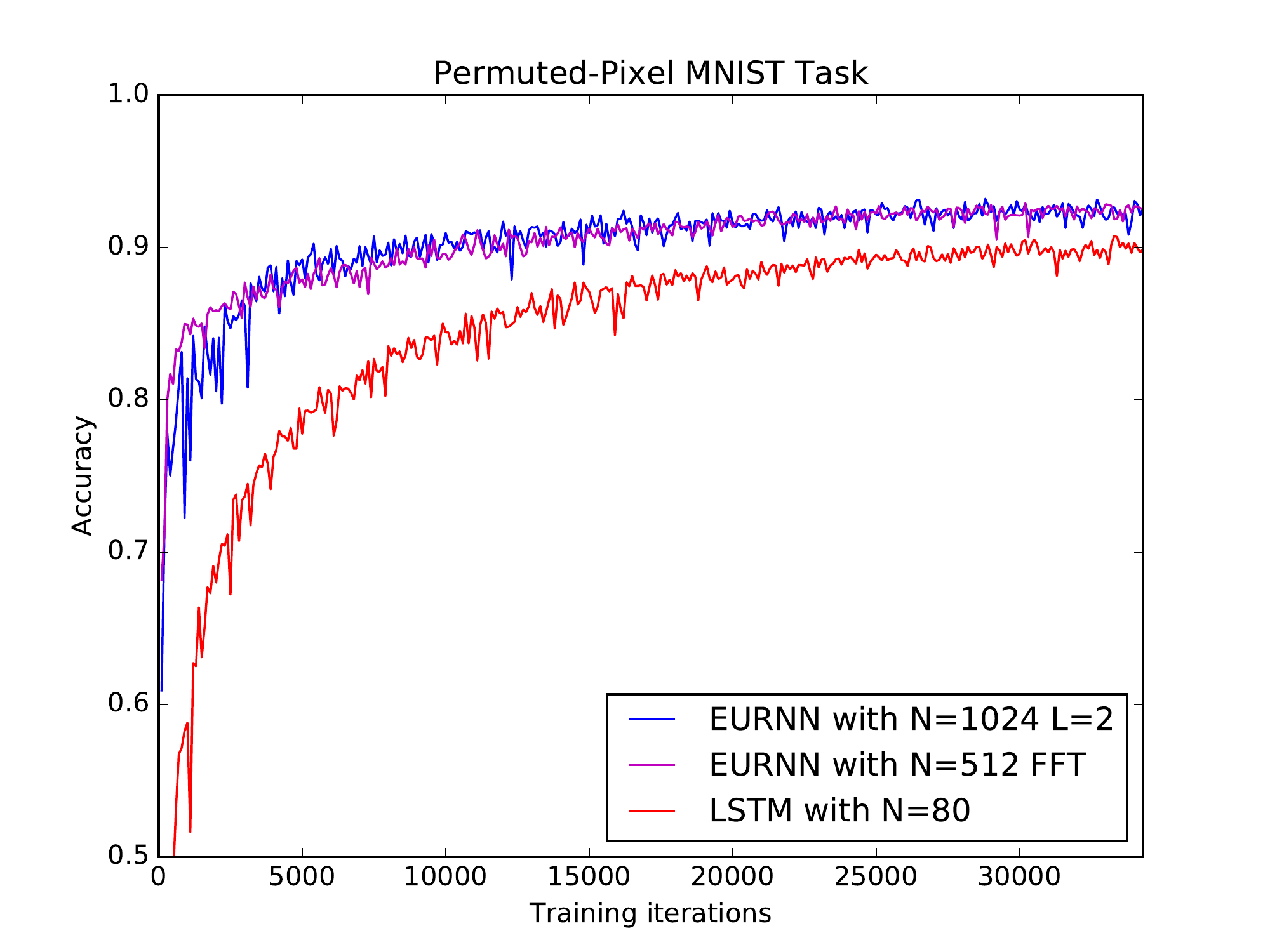}
\caption{Pixel-permuted MNIST performance on the validation dataset.}
\label{fig:MNIST}
\end{figure} 

As shown in Fig.~\ref{fig:MNIST}, EURNN significantly outperforms LSTM with the same number of parameters. It learns faster, in fewer iteration steps, and converges to a higher classification accuracy. In addition, the EURNN reaches a similar accuracy with fewer parameters. In Table.~\ref{tab:mnist}, we compare the performance of different RNN models on this task.

\subsection{Speech Prediction on TIMIT dataset}

\begin{table*}[t!]
\centering
\caption{Speech Prediction Task result. EURNN corresponds to our algorithm, projective URNN corresponds to algorithm presented in \cite{wisdom2016full}, URNN corresponds to the algorithm presented in \cite{arjovsky2015unitary}.}
\vspace{0.1in}
\begin{tabular}{ccccc}
\hline
Model           & hidden size & number of  & MSE &  MSE \\
				& (capacity) & parameters & (validation) & (test)\\
\hline
LSTM					& 64 & 33k & 71.4 & 66.0  \\
LSTM					& 128 & 98k & 55.3 & 54.5  \\
EURNN (tunable style)     & 128 (2) & 33k & 63.3 & 63.3  \\
EURNN (tunable style)     & 128 (32) & 35k & 52.3 & 52.7 \\
EURNN (tunable style)     & 128 (128) & 41k & \bf{51.8} & \bf{51.9} \\
EURNN (FFT style)   & 128 (FFT) & 34k & 52.3 & 52.4 \\
\hline
\end{tabular}
\label{tab:timit}
\end{table*}

\begin{figure*}[t!]
\label{fig:timit}
\centering
\includegraphics[width=7in]{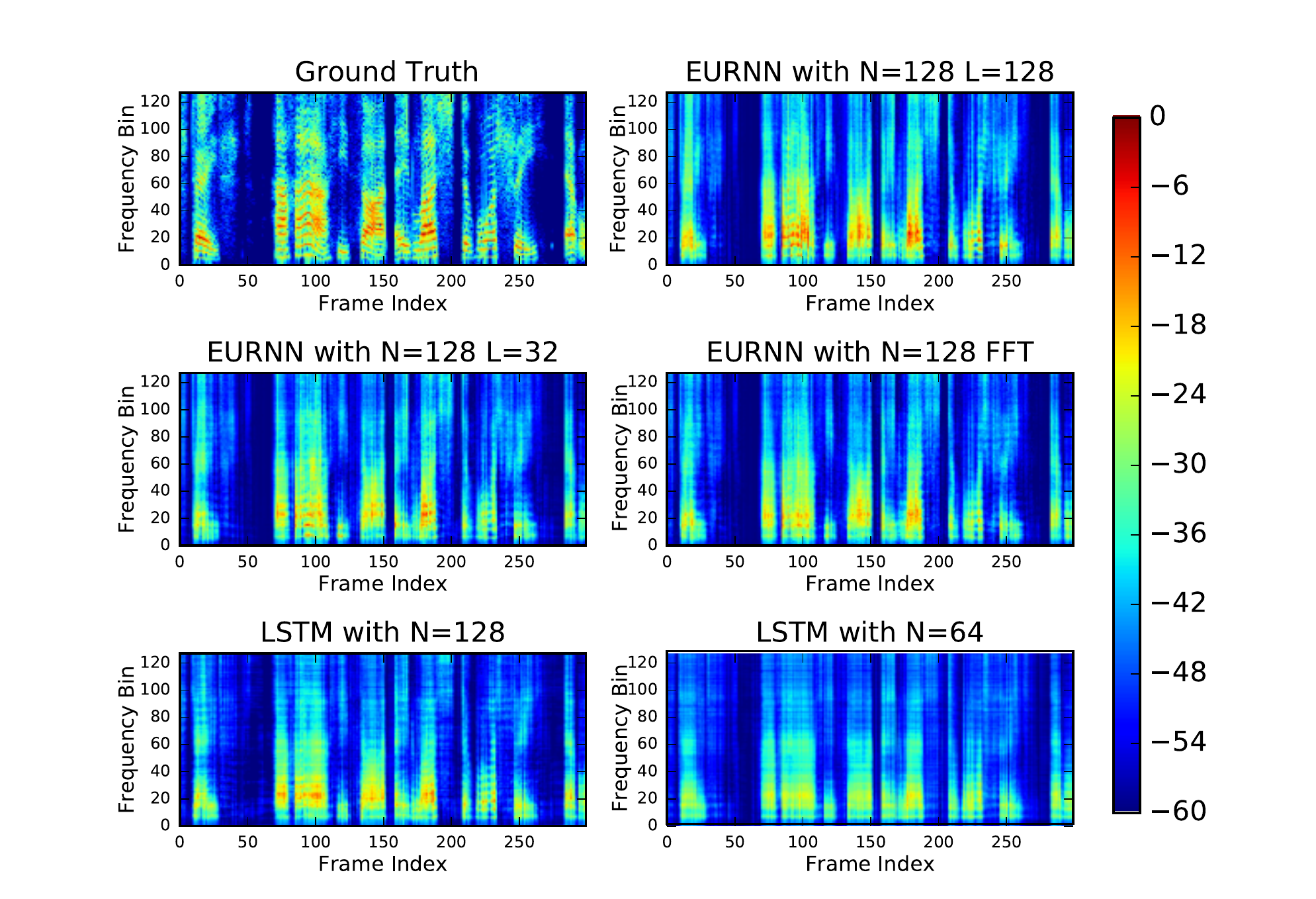}
\caption{Example spectrograms of ground truth and RNN prediction results from evaluation sets.}
\label{fig:TIMIT}
\end{figure*}

We also apply our EURNN to real-world speech prediction task and compare its performance to LSTM. The main task we consider is predicting the log-magnitude of future frames of a short-time Fourier transform (STFT) \cite{wisdom2016full,SEJDIC2009153}. We use the TIMIT dataset \cite{garofolo1993darpa} sampled at 8 kHz. The audio \textbf{.wav} file is initially diced into different time frames (all frames have the same duration referring to the Hann analysis window below). The audio amplitude  in each frame is then Fourier transformed into the frequency domain. The log-magnitude of the Fourier amplitude is normalized and used as the data for training/testing each model. In our STFT operation we uses a Hann analysis window of 256 samples (32 milliseconds) and a window hop of 128 samples (16 milliseconds). The frame prediction task is as follows: given all the log-magnitudes of STFT frames up to time $t$, predict the log-magnitude of the STFT frame at time $t + 1$ that has the minimum mean square error (MSE). We use a training set with 2400 utterances, a validation set of 600 utterances and an evaluation set of 1000 utterances. The training, validation, and evaluation sets have distinct speakers. We trained all RNNs for with the same batch size 32 using RMSProp optimization with a learning rate of 0.001, a momentum of 0.9 and a decay rate of 0.1.

The results are given in Table.~\ref{tab:timit}, in terms of the mean-squared error (MSE) loss function. Figure.~\ref{fig:TIMIT} shows prediction examples from the three types of networks, illustrating how EURNNs generally perform better than LSTMs. Furthermore, in this particular task, full-capacity EURNNs outperform small capacity EURNNs and FFT-style EURNNs.






\section{Conclusion}

We have presented a method for implementing an Efficient Unitary Neural Network (EUNN) whose computational cost is merely $\mathcal{O}(1)$ per parameter, which is $\mathcal{O}(\log N)$ more efficient than the other methods discussed above. 
It significantly outperforms existing RNN architectures on the standard Copying Task, and the pixel-permuted MNIST Task using a comparable parameter count, hence demonstrating the highest recorded ability to memorize sequential information over long time periods. 

It also performs well on real tasks such as speech prediction, outperforming an LSTM on TIMIT data speech prediction.



We want to emphasize the generality and tunability of our method. The ordering of the rotation matrices we presented in Fig.~\ref{fig:decomp} are merely two of many possibilities; we used it simply as a concrete example. Other ordering options that can result in spanning the full unitary matrix space can be used for our algorithm as well, with identical speed and memory performance. This tunability of the span of the unitary space and, correspondingly, the total number of parameters makes it possible to use different capacities for different tasks, thus opening the way to an optimal performance of the EUNN.
For example, as we have shown, a small subspace of the full unitary space is preferable for the copying task, whereas the MNIST task and TIMIT task are better performed by EUNN covering a considerably larger unitary space. Finally, we note that our method remains applicable even if the unitary matrix is decomposed into a different product of matrices (Eq. 12).

This powerful and robust unitary RNN architecture also might be promising for natural language processing because of its ability to efficiently handle tasks with long-term correlation and very high dimensionality. 



\section*{Acknowledgment}
We thank Hugo Larochelle and Yoshua Bengio for helpful discussions and comments. 

This work was partially supported by the Army Research Office through the Institute for Soldier Nanotechnologies under contract W911NF-13-D0001, the National Science Foundation under Grant No. CCF-1640012 and the Rothberg Family Fund for Cognitive Science.

\bibliography{citations}
\bibliographystyle{icml2017}

\end{document}